\pdfoutput=1

\documentclass[11pt]{article}

\usepackage[preprint]{acl}

\usepackage{times}
\usepackage{latexsym}

\usepackage[T1]{fontenc}

\usepackage[utf8]{inputenc}

\usepackage{microtype}

\usepackage{inconsolata}

\usepackage{graphicx}

\usepackage{amsfonts}
\usepackage[most]{tcolorbox}
\usepackage{cleveref}
\usepackage{float}
\usepackage{booktabs}
\usepackage{multirow}
\usepackage{enumitem}
\usepackage{algorithm}
\usepackage{algpseudocode}
\usepackage{placeins}
\usepackage{listings}
\title{Do Large Language Models Truly Grasp Addition?\\A Rule-Focused Diagnostic Using Two-Integer Arithmetic}

\author{
    Yang Yan$^{1,2}$, 
    Yu Lu$^{2}$, 
    Renjun Xu$^{1,}$\thanks{Corresponding Authors.},
    Zhenzhong Lan$^{2,}$\footnotemark[1]\\
    $^{1}$ Zhejiang University \\ 
    $^{2}$ School of Engineering, Westlake University \\
    \texttt{\{yan.yang,rux\}@zju.edu.cn} \\
    \texttt{\{yanyang,luyu,lanzhenzhong\}@westlake.edu.cn}\\
}

\begin{document}
\maketitle
\begin{abstract}

Large language models (LLMs) achieve impressive results on advanced mathematics benchmarks but sometimes fail on basic arithmetic tasks, raising the question of whether they have \emph{truly grasped} fundamental arithmetic rules or are merely relying on pattern matching. To unravel this issue, we systematically probe LLMs’ understanding of two-integer addition ($0$ to $2^{64}$) by testing three crucial properties: commutativity ($A+B=B+A$), representation invariance via symbolic remapping (e.g., $7\mapsto\textsc{Y}$), and consistent accuracy scaling with operand length. Our evaluation of 12 leading LLMs reveals a stark disconnect: while models achieve high numeric accuracy (73.8–99.8\%), they systematically fail these diagnostics. Specifically, accuracy plummets to $\le 7.5$\% with symbolic inputs, commutativity is violated in up to 20\% of cases, and accuracy scaling is non-monotonic. Interventions further expose this pattern-matching reliance: explicitly providing rules degrades performance by 29.49\%, while prompting for explanations before answering merely maintains baseline accuracy. These findings demonstrate that current LLMs address elementary addition via pattern matching, not robust rule induction, motivating new diagnostic benchmarks and innovations in model architecture and training to cultivate genuine mathematical reasoning. we release both our diagnostic dataset and the code for dataset generation at \url{https://github.com/kuri-leo/llm-arithmetic-diagnostic}.
\end{abstract}

\begin{figure}[t]
\includegraphics[width=\columnwidth]{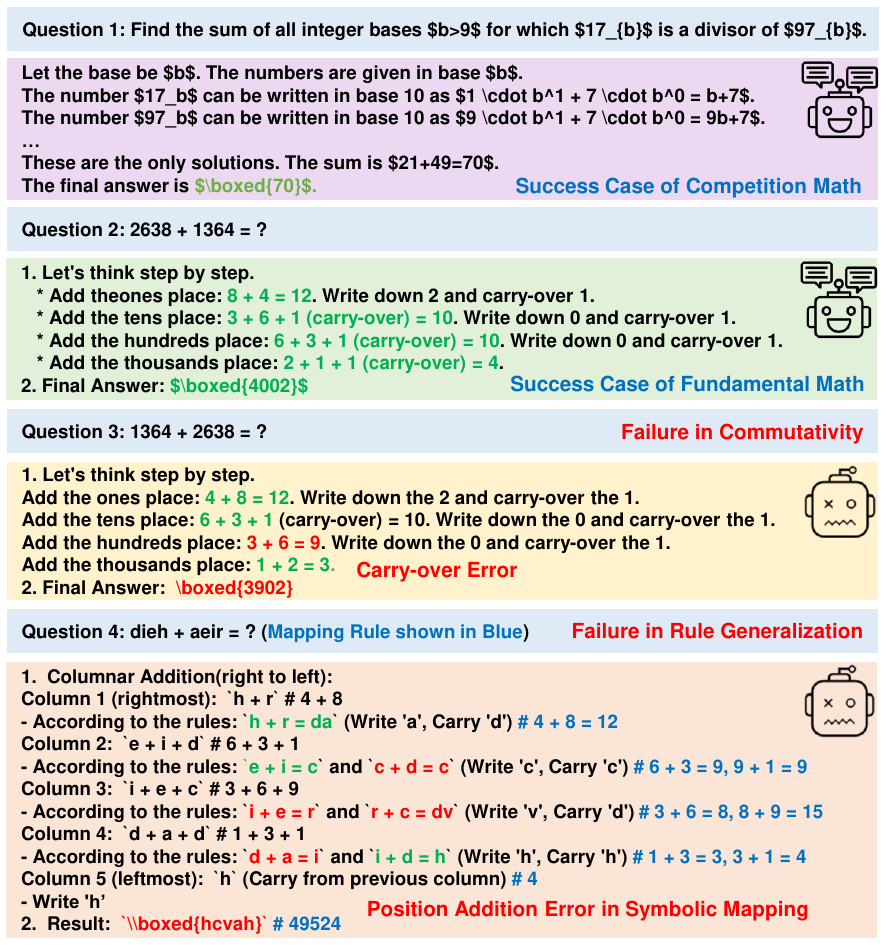}
\caption{ \textbf{Illustration of LLM Paradox: LLMs excel at complex math but falter on basic addition, raising the question of whether they grasp rules or merely reproduce patterns.} ``True grasp'' implies consistent performance and adherence to mathematical properties (e.g., commutativity, representation invariance) under novel conditions. This study probes LLMs' comprehension of elementary two-integer addition (RQ1) and the factors that modulate it (RQ2). Findings suggest that LLMs rely on token-level heuristics rather than rule abstraction.}
\label{fig:experiments}
\end{figure}

\section{Introduction}

\paragraph{Why basic arithmetic still matters.}
Large Language Models (LLMs) have demonstrated impressive, often graduate-level, performance on sophisticated mathematics benchmarks~\citep{openai2024o1,deepseekai2025deepseekr1}. This success, however, is frequently contrasted with surprising failures on \emph{elementary} operations when inputs are minimally perturbed, such as reordering operands, substituting digits with symbols, or rephrasing problems~\citep{li2024gsmplus,mirzadeh2025gsmsymbolic}. This stark contradiction fuels a critical question at the heart of understanding LLM capabilities:
\begin{quote}
    \textbf{Do LLMs truly grasp arithmetic \emph{rules}, or do they primarily reproduce familiar token patterns learned from vast datasets?}
\label{eq:question}
\end{quote}
\noindent Here, we define \emph{rules} as generalizable algorithms that apply consistently across all valid inputs, and \emph{true grasp} as the ability to maintain performance and adhere to mathematical properties even under novel conditions, reflecting an understanding of the principles themselves.

\paragraph{Diagnostic gap in existing benchmarks.}
Popular benchmarks such as GSM8K, MATH-500, and Humanity's Last Exam emphasize final-answer accuracy on multi-step word problems~\citep{cobbe2021training,lightman2024lets,phan2025humanitysexam}.
Because many sub-steps are unobserved, these benchmarks cannot determine whether success arises from rule learning or from distribution-specific heuristics.
Recent robustness probes confirm this concern, but still involve problems whose complexity obscures \textbf{which precise rule is violated}~\citep{li2024gsmplus,mirzadeh2025gsmsymbolic}.
A targeted diagnostic capable of isolating algorithmic execution from linguistic complexity is therefore required.

\paragraph{Our diagnostic lens: two-integer addition.}
We propose elementary two-integer addition as a controlled probe of rule learning (\Cref{fig:experiments}).
The task isolates a single algorithm with two components, digit-wise addition and carry propagation, and removes linguistic confounds.
Any system that genuinely implements this algorithm must satisfy three observable properties:
(i) \textbf{digit-scaling consistency}, meaning accuracy should be non-increasing with operand length;
(ii) \textbf{representation invariance}, meaning performance should be stable under any bijective digit-to-symbol remapping; and
(iii) \textbf{algebraic integrity}, meaning commutativity holds for every operand pair.

\paragraph{Empirical Findings.}
Our empirical investigation, centered on these properties, reveals that contemporary LLMs predominantly rely on pattern matching rather than exhibiting a robust, rule-based understanding of elementary addition. Key findings highlight significant violations of digit-scaling consistency: numeric accuracy often shows an erratic 'drop-rebound' pattern, such as declining for 4–6 digit operands, then improving for 8–10 digits, instead of the expected monotonic degradation with increasing operand length. LLMs also frequently fail to uphold algebraic integrity; commutativity ($A+B \neq B+A$) is systematically violated in thousands of instances across various models, with some models failing this property in up to 20\% of problem pairs. Moreover, performance dramatically collapses under symbolic representation; models achieving over 99\% numeric accuracy, like Claude-3.5-Sonnet at 99.81\%, experience a performance drop to as low as 7.51\% when standard digits are replaced by novel symbols. Interventions such as providing explicit rules via prompting often counterintuitively degrade performance, sometimes by more than 50\% relative to zero-shot accuracy. Finally, fine-tuning experiments underscore a persistent tension. Task-specific supervised fine-tuning (SFT) significantly boosts numeric accuracy but typically fails to generalize this improvement to symbolic tasks. In contrast, reinforcement learning (RL) based methods show somewhat better symbolic transfer, although often without matching SFT's peak numeric performance. These outcomes indicate that LLMs' arithmetic competence is often tied to learned surface token patterns, rather than an internalized, abstract grasp of mathematical rules.

\paragraph{Contributions.}
Our main contributions are:
\begin{enumerate}[leftmargin=0pt, itemindent=1.3em, itemsep=0pt, parsep=0pt, topsep=0pt, partopsep=0pt]
	\item \textbf{Diagnostic Methodology:} We introduce a diagnostic methodology centered on two-integer addition, using notation invariance, digit-scaling consistency, and algebraic integrity as key criteria to differentiate genuine rule learning from superficial pattern matching in LLMs.
	\item \textbf{Empirical Findings:} Through extensive experiments, we demonstrate that current LLMs systematically fail these diagnostic tests, exhibiting significant performance drops with symbolic inputs, erratic scaling, and commutativity violations. Furthermore, interventions like explicit rule provision often impair performance, while fine-tuning highlights a persistent preference for pattern memorization over rule abstraction.
	\item \textbf{Implications for LLM Evaluation and Development:} These findings reveal a core limitation in LLMs' compositional generalization for elementary arithmetic. This suggests that success on complex benchmarks may mask deficiencies in foundational reasoning, underscoring the need for new evaluation approaches and model architectures to foster genuine mathematical understanding.
\end{enumerate}

\section{Related Work}

\begin{itemize}[leftmargin=0pt,itemindent=0em,itemsep=0pt]
\item[] \textbf{Benchmark progress and limits.}
Leaderboard-oriented benchmarks have propelled LLMs toward increasingly impressive performance on complex reasoning tasks, ranging from general knowledge assessments to graduate-level mathematics~\citep{hendrycks2021measuring, cobbe2021training, aime2024, phan2025humanitysexam}. However, these suites prioritize final answers over the underlying generalizable rules that generate them. Since each problem combines multiple subtasks, high accuracy can be achieved through localized pattern matching, which can evade detection by aggregate metrics. Thus, the fundamental question persists, and our work specifically addresses this uncertainty.

\item[] \textbf{Robustness analyses that reveal surface dependence.}
This reliance on familiar tokens is a specific instance of a broader challenge in domain adaptation and out-of-distribution (OOD) generalization~\citep{yuan2024revisiting}. Several studies probe models with minimal input perturbations, revealing a critical dependence on surface patterns. In mathematical reasoning, models that excel on standard benchmarks show significant performance drops when numerical values are changed, irrelevant clauses are added, or problems are rephrased~\citep{li2024gsmplus, mirzadeh2025gsmsymbolic}. This fragility suggests models replicate shallow heuristics rather than executing robust algorithms, a limitation observed even in controlled, non-linguistic puzzles where reasoning capabilities collapse with rising complexity~\citep{zhang2024illusion}. This mirrors findings in other domains where models latch onto superficial content features rather than generalizable structural rules~\citep{roussinov2025controlling}. Similarly, altering numeral formats or retokenizing inputs also consistently reduces accuracy~\citep{zhou2024scaling,zhong2024achieving,zeng2024mrgsm8k}. Although specialized embeddings can recover some performance~\citep{mcleish2024transformers}, these fixes improve specific surface forms and do not demonstrate rule abstraction. Our notation-remapping experiments extend this line of inquiry by isolating the addition algorithm from every other linguistic cue, providing a focused test of OOD generalization for a fundamental rule.

\item[] \textbf{Mechanistic studies of arithmetic circuits.}
Neuron-level inspections report units that store partial carries, as well as heuristics that fail outside the training range~\citep{qiu2024dissecting,nikankin2024arithmetic}.  
Grokking phenomena illustrate that models can memorize before they generalize, and sometimes never reach full rule induction~\citep{power2022grokking}.  
Instruction-tuning and in-context exemplars can elicit temporary algebraic behavior, yet systematic transfer remains narrow~\citep{chang2024unraveling,gorceix2024learning,chen2024states,deng2024language}.  
These findings motivate a diagnostic that tests rule mastery directly rather than inferring it from indirect proxies.

\end{itemize}

We contribute such a diagnostic by focusing on two-integer addition, a task whose solution requires commutativity and notation invariance but avoids confounds from multi-step language understanding. By evaluating models against these minimal yet stringent criteria, we close the empirical gap left by prior robustness and mechanistic studies and provide a concrete baseline for future architectural and training advances.

\begin{figure*}[t]
\centering
\includegraphics[width=\linewidth]{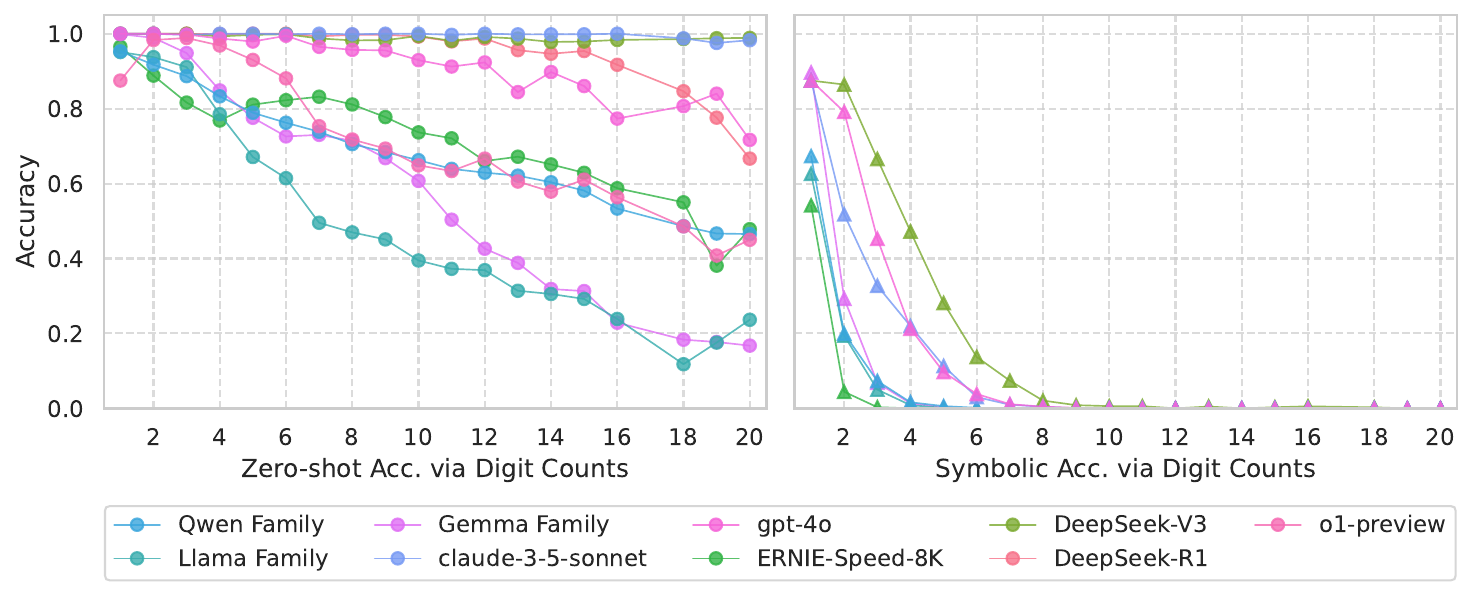}
\caption{\textbf{Performance Degradation Patterns in Zero-shot vs. Symbolic Addition.} While LLMs achieve high accuracy on standard numerical addition (left), their non-monotonic performance curve suggests brittle pattern matching rather than true algorithmic reasoning. In contrast, symbolic addition tests (right) reveal systematic degradation with increasing digit count. This stark contrast between numerical and symbolic performance suggests LLMs rely heavily on memorized patterns rather than learned arithmetic principles.}
\label{fig:zero_shot_symbolic_digit_count}
\end{figure*}

\section{Methodology}
\label{sec:method}

\subsection{Background and Motivation}
\label{sec:prelim}
LLMs predominantly employ auto-regressive generation to produce responses. Given a question \(Q\), an LLM samples an answer sequence \(A = (T_1, \dots, T_N)\) from a learned probability distribution of pre-training data:
\begin{equation}
A \sim P(A \mid Q) = \prod_{i=1}^{N} P\left(T_i \mid T_{<i}, T\right),
\label{eq:autoregressive}
\end{equation}
Ideally, this learned distribution $P(A \mid Q)$ should approximate a true underlying distribution $P^*(A \mid Q)$ that reflects genuine comprehension, or \emph{true grasp}, of the principles needed to answer $Q$. Such genuine comprehension would manifest as consistent behavior governed by the task's abstract properties, not surface-level statistics. For example, the resulting distribution should be invariant to superficial changes like re-ordering operands or remapping notation, and its performance should scale predictably with task complexity. However, empirical evidence indicates that LLMs often violate these properties: high accuracy on complex benchmarks like GSM8K can significantly decrease with minor input perturbations, such as paraphrasing or digit substitution, or on simpler arithmetic tasks \citep{li2024gsmplus, mirzadeh2025gsmsymbolic, zhou2024scaling}. This discrepancy between the learned $P(A \mid Q)$ and the ideal $P^*(A \mid Q)$ suggests a reliance on surface-level pattern matching rather than internalized, rule-governed reasoning.

While the complexity of word-problem benchmarks obscures precise failure attribution, an ideal diagnostic task should isolate a single, verifiable algorithm to test a model's ability for procedural execution, which is a foundational capability distinct from the semantic or associative reasoning probed in other robustness studies \citep{berglund2024the, alhamoud2025vision, nguyen2024counterfactuals}. Such a task allows us to probe not just what a model knows, but how it computes. Two-integer addition, a classic micro-benchmark for algorithmic generalization \citep{saxton2019analysing}, meets these criteria, offering a simple algorithm and an input domain too vast for memorization.

Our methodology thus rigorously probes LLMs' grasp of elementary addition, addressing:
\begin{itemize}[leftmargin=0pt,itemindent=0em,itemsep=0pt]
	\item[] \textbf{RQ1:} Do LLMs satisfy key markers of rule-based addition: notation invariance, digit-scaling consistency, and algebraic integrity?
	\item[] \textbf{RQ2:} Can prompt-level or parameter-level interventions bridge the gap between pattern recall and rule induction?
\end{itemize}
To answer these questions, we construct a dataset of two-integer addition problems and evaluate LLMs against three diagnostic properties: digit-scaling consistency, representation invariance, and algebraic integrity. We also explore the effects of explicit rule provision and prompting strategies on model performance in following sections.

\subsection{Diagnostic Task: Two-Integer Addition}
\label{sec:task}

Two-integer addition serves as our diagnostic task. It isolates a simple but complete arithmetic algorithm (core components: per-digit addition, carry propagation), requires generalization beyond memorization, and permits unambiguous verification. To operationally distinguish between genuine rule induction and superficial pattern matching, \textbf{we derive three fundamental properties that any model demonstrating a true grasp of the addition algorithm must consistently exhibit}:

\begin{enumerate}[leftmargin=0pt, itemindent=1.3em, itemsep=0pt, parsep=0pt, topsep=0pt, partopsep=0pt]
	\item \textbf{Digit-scaling consistency:} For a model applying a consistent algorithm, accuracy should be non-increasing (i.e., stable or monotonically decreasing) as operand length increases, reflecting cumulative error potential. Any deviation from this, particularly a non-monotonic 'drop-rebound' pattern, indicates reliance on length-specific heuristics rather than a scalable rule.
	\item \textbf{Algebraic integrity:} Fundamental algebraic properties, such as commutativity ($A+B=B+A$), must be consistently upheld. A model with a genuine grasp should yield $P(S\mid\text{Query}(A+B)) \approx P(S\mid\text{Query}(B+A))$. Systematic violations of such properties provide direct evidence against true rule grasp.
	\item \textbf{Representation invariance:} Performance must remain robust when standard digits are bijectively mapped to novel, arbitrary symbols. A significant degradation in performance under such mapping implies that the learned distribution $P(A\mid Q)$ is overfit to surface token patterns, rather than grasping the abstract rule captured by $P^*(A\mid Q)$.
\end{enumerate}

These three properties serve as direct litmus tests for distinguishing genuine rule induction from superficial pattern matching.

\subsection{Dataset Construction}
\label{sec:data}

To systematically evaluate LLMs against our diagnostic properties, we construct a comprehensive dataset of 100,000 unique two-integer addition problems. The operands A and B are sampled from the range $[0, 2^{64}-1)$. We structure the dataset generation in three phases to ensure thorough coverage and facilitate targeted analyses:

\begin{enumerate}[leftmargin=0pt, itemindent=1.3em, itemsep=0pt, parsep=0pt, topsep=0pt, partopsep=0pt]
	\item \textbf{Phase 1 (Baseline):} All two-digit addition pairs (0-99) for fundamental assessment.
	\item \textbf{Phase 2 (Digit Scaling):} Uniform sampling of same-length operand pairs (3-20 digits), enabling assessment of length-dependent performance.
	\item \textbf{Phase 3 (Large Numbers):} Additional samples from \(2^{49}\) to \(2^{64}\), testing robustness.
\end{enumerate}

To directly assess our diagnostic properties, the dataset incorporates specific structural features. Algebraic integrity, particularly commutativity, is systematically evaluated by including the commuted counterpart (B,A) for every generated ordered pair (A,B). For testing representation invariance, we define ten independent bijective digit-to-symbol mappings (e.g., $0 \mapsto \textsc{u}, \dots, 9 \mapsto \textsc{c}$) and apply them to a designated subset of numerical problems to create corresponding symbolic variants. The dataset is subsequently split into training (80\%), validation (10\%), and test (10\%) portions. To ensure efficiency, evaluations of proprietary, reasoning, and fine-tuned models are restricted to this test set. This comprehensive dataset design facilitates a thorough assessment of both rule-learning criteria and intervention effects through systematic property verification.

\section{Experiments}
\label{sec:experiment}
To empirically test our central hypothesis, that LLMs rely on pattern matching over rule induction for elementary addition, and to validate our diagnostic methodology (\Cref{sec:method}), we conducted experiments addressing the two RQs:

\begin{itemize}[leftmargin=0pt, itemindent=1.3em, itemsep=0pt, parsep=0pt, topsep=0pt, partopsep=0pt]
    \item \textbf{RQ1: Do LLMs satisfy key markers of rule-based addition?} To answer this, we evaluate a diverse set of LLMs against the three core diagnostic properties derived from our methodology: digit-scaling consistency, algebraic integrity (specifically commutativity), and notation invariance.
    \item \textbf{RQ2: Can prompt-level or parameter-level interventions bridge the gap between pattern recall and rule induction?} Here, we probe how LLM performance on addition is modulated by explicit rule provision, self-explanation prompting, and task-specific fine-tuning.
\end{itemize}

\noindent We evaluated a diverse range of contemporary LLMs, including open-source families (e.g., \texttt{Llama}, \texttt{Qwen}) and proprietary LLMs (e.g., \texttt{GPT-4}, \texttt{Gemini} series). Due to API costs, evaluations for some proprietary models were limited; full details are in \Cref{app:experimental_setup}. These experiments provide empirical evidence for our claims about LLM arithmetic capabilities.

\subsection{RQ1: Do LLMs Truly Grasp Addition?}
\label{sec:rq1}

\begin{table}
\centering
\resizebox{\linewidth}{!}{
\begin{tabular}{lrrrrrr}
\toprule
\multicolumn{1}{l}{\textbf{Task Type}} & \multicolumn{2}{r}{zero-shot} & symbolic & \multicolumn{2}{r}{zero-shot} & symbolic \\
\multicolumn{1}{l}{\textbf{Temp}} & 0.0 & 0.7 & 0.7 & 0.0 & 0.7 & 0.7 \\
\multicolumn{1}{l}{\textbf{Violation Threshold \#}} & 5 & 5 & 5 & 10 & 10 & 10 \\
\midrule
Llama3-8B-it & 11918 & 1678 & 41 & 4783 & 1 & - \\
Llama3-70B-it & - & 5402 & 506 & - & 432 & 20 \\
Llama3.1-8B-it & 13324 & 3232 & 49 & 4546 & 6 & - \\
Llama3.1-70B-it & - & 4546 & 602 & - & 253 & 22 \\
Llama3.3-70B-it & - & 5086 & 1122 & - & 1771 & 81 \\
Qwen2.5-7B-it & 7442 & 3961 & 302 & 3402 & 151 & 29 \\
Qwen2.5-72B-it & - & 812 & 745 & - & 25 & 10 \\
\bottomrule
\end{tabular}
}
\caption{\textbf{Commutativity Violations Statistics.}
Each entry reports the number of (A,B) pairs, out of 50,000, where the model correctly computes $A{+}B$ but fails on $B{+}A$; lower counts indicate better performance.
Columns correspond to different decoding temperatures (`Temp') and minimum thresholds (5 or 10 identical successes out of 10 samples in total). A `-' indicates that an evaluation was not run due to cost, as detailed in \Cref{app:experimental_setup}.}
\label{tab:commutative_property_failure_cases}
\end{table}

To determine if LLMs have truly grasp the addition rules, as opposed to merely mimicking patterns, we evaluated their performance against the three properties. Failure to satisfy these properties would serve as strong evidence of a superficial understanding, reliant on surface-level heuristics rather than genuine rule learning. We present the empirical findings for each diagnostic in turn.

\paragraph{Violation of Digit-Scaling Consistency.}
Our first diagnostic, digit-scaling consistency, posits that accuracy should be non-increasing with operand length for a system that has internalized an iterative algorithm like addition. However, LLMs frequently violate this principle. As shown in \Cref{fig:zero_shot_symbolic_digit_count} (left panel), many models exhibit a non-monotonic `drop-rebound' accuracy curve on numeric inputs: performance initially declines for 4–6 digit operands, then unexpectedly improves for 8–10 digits, before declining again. This erratic scaling suggests that the learned distribution $P(A \mid Q)$ for numeric addition deviates from the ideal $P^*(A \mid Q)$ defined in \Cref{sec:prelim}, indicating reliance on length-specific heuristics or memorized fragments rather than a general, scalable rule. In contrast, when inputs are symbolic (\Cref{fig:zero_shot_symbolic_digit_count}, right panel), accuracy, while substantially lower, tends to decline more monotonically with increasing digit count. This latter pattern, paradoxically more aligned with algorithmic processing under stress, reinforces the interpretation that high performance on standard numeric inputs reflects pattern matching rather than the robust rule application outlined by our criteria in \Cref{sec:task}.

\paragraph{Failure to Uphold Algebraic Integrity.}
We assessed algebraic integrity by testing adherence to commutativity, a fundamental property of addition. A violation occurred if a model correctly computed $A+B$ but not $B+A$ (or vice versa) in at least 5 (or all 10) of 10 stochastic decodes per problem, at temperatures `Temp'=0.0 and `Temp'=0.7. The results (\Cref{tab:commutative_property_failure_cases}) reveal frequent and systematic commutativity failures. For instance, some 7-8B Llama and Qwen models processing numerical inputs violated commutativity in approximately 20\% of 50,000 problem pairs (when failing $\ge 5/10$ decodes) and in 8.48\% (4,243 pairs) when failing all 10 decodes. These inconsistencies persisted even at `Temp'=0.7, where 0.104\% of pairs (52 instances) showed such persistent violations across all decodes. Conversely, on symbolic tasks, Qwen2.5-7B showed such consistent (10/10) violations in only 0.058\% of pairs (29 instances), while Llama models exhibited none. Moreover, the observation that larger Llama models were sometimes more prone to these violations challenges the notion that increased model scale inherently confers deeper arithmetic understanding. Such widespread and systematic failures strongly suggest inherent deficiencies in the models' understanding of the addition algorithm, rather than random noise.

\paragraph{Breakdown under Tests of Notation Invariance.}
Our third diagnostic, notation invariance, assesses if LLM performance on addition remains stable when standard digits are bijectively mapped to novel symbols, a property expected if the abstract addition algorithm is truly internalized. As detailed in \Cref{tab:zero_shot_symbolic}, all tested models dramatically failed this test. Even those with near-perfect accuracy on numeric inputs (e.g., 99.81\% for \texttt{Claude-3.5-Sonnet}) saw performance collapse on symbolic equivalents, to as low as 7.51\%. This failure extended to fundamental components like positional addition and carry-over sub-tasks once familiar digit patterns were absent. Such pronounced inability to generalize to novel symbols strongly indicates that current LLMs rely on recognizing and reproducing patterns tied to standard decimal representations, rather than having learned an abstract, symbol-agnostic addition rule.

\paragraph{Generalization to Other Operations and Semantic Contexts.} To validate the generalizability of our findings, we extended our methodology to subtraction and multiplication. As shown in \Cref{tab:other_operations}, LLMs exhibit similar performance collapses, confirming their struggles extend beyond addition. Furthermore, to address the artificiality of the task, we embedded problems into natural language templates following \citet{mirzadeh2025gsmsymbolic}. While performance slightly improves in this `semantic' context, the same fundamental failures, including the non-monotonic 'drop-rebound' pattern, persist (see Appendix for full results). This indicates our findings reflect a broader limitation and are not merely an artifact of a single task.

\begin{table}[t]
\centering
\resizebox{\linewidth}{!}{
\begin{tabular}{lrrrrrr}
\toprule
& \multicolumn{2}{c}{Overall Acc.} & \multicolumn{2}{c}{Position Add Acc.} & \multicolumn{2}{c}{Carry-over Acc.} \\
Task Type & ZS & S & ZS & S & ZS & S \\
\midrule
Gemini2.0-pro-exp & 94.88 & 14.21 & 69.52 & 4.19 & 77.36 & 7.07 \\
Claude-3.5-Sonnet & 99.81 & 7.51 & 81.78 & 3.19 & 90.28 & 6.92 \\
GPT-4o & 93.39 & 9.59 & 76.12 & 3.79 & 79.55 & 6.73 \\
DeepSeek-V3 & 98.92 & 16.14 & 78.55 & 11.98 & 81.14 & 15.23 \\
\midrule
Gemma2-9b-it & 66.34 & 1.45 & 58.52 & 0.34 & 60.44 & 0.44 \\
Gemma2-27b-it & 83.65 & 2.62 & 74.77 & 0.91 & 76.68 & 0.91 \\
\cline{2-7}\\[-5pt]
Llama3.1-8B-it & 43.34 & 0.57 & 20.38 & 0.10 & 21.96 & 0.25 \\
Llama3.1-70B-it & 72.58 & 2.51 & 60.13 & 0.52 & 61.05 & 1.33 \\
\cline{2-7}\\[-5pt]
Qwen2.5-7B-it & 83.00 & 0.58 & 71.39 & 0.11 & 74.49 & 0.13 \\
Qwen2.5-72B-it & 96.07 & 5.97 & 88.19 & 2.09 & 89.78 & 4.12 \\
\bottomrule
\end{tabular}
}
\caption{\textbf{Accuracy on elementary two-integer addition.}
\textbf{ZS} = zero-shot numeric form; \textbf{S} = Symbolic form (bijective digit-to-symbol mapping). The full table including performance degradation ($\Delta$) is in the Appendix.
}
\label{tab:zero_shot_symbolic}
\end{table}

Collectively, the evidence from these three diagnostic tests converges on a clear and consistent conclusion: despite often achieving high accuracy on standard numeric addition problems, contemporary LLMs do not demonstrate a robust, rule-based understanding of this elementary operation. Their competence appears tightly coupled to familiar surface token patterns and specific operand lengths, and it degrades systematically when these patterns are disrupted or when fundamental algebraic properties are rigorously tested. This pattern of behavior strongly indicates a primary reliance on pattern matching rather than genuine rule induction for performing elementary addition.

Having established these fundamental deficiencies in LLMs' grasp of basic addition, we next investigate factors that might modulate this understanding in \Cref{sec:rq2}.

\begin{table}[t]
\centering
\resizebox{\linewidth}{!}{
\begin{tabular}{llrr}
\toprule
Task & Task Type & Llama3.1-8B-it & Qwen2.5-7B-it \\
\midrule
\multirow[c]{2}{*}{Add} & symbolic & 0.64 & 0.58 \\
& zero-shot & 43.43 & 83.03 \\
\cline{1-4}
\multirow[c]{2}{*}{Multi} & symbolic & 0.01 & 0.04 \\
& zero-shot & 9.92 & 17.29 \\
\cline{1-4}
\multirow[c]{2}{*}{Sub} & symbolic & 0.02 & 0.01 \\
& zero-shot & 18.39 & 43.88 \\
\bottomrule
\end{tabular}
}
\caption{ \textbf{Performance on Other Arithmetic Operations.} Building on observed difficulties with addition, we evaluated subtraction and multiplication. LLMs performed poorly on symbolic representations of these operations, with low zero-shot accuracy. This suggests their struggles extend to these more complex operations and their symbolic forms.}
\label{tab:other_operations}
\end{table}

\begin{figure*}[t]
\centering
\includegraphics[width=\linewidth]{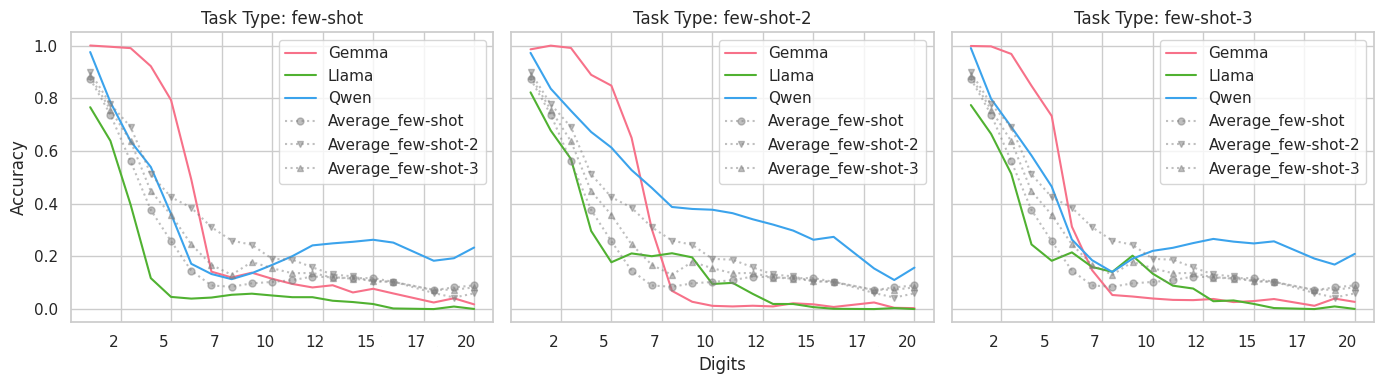}
\caption{ \textbf{Few-Shot Performance with Explicit Rule Provision.} Explicit rule provision leads to a significant drop in performance compared to zero-shot, contradicting the expected improvement. The \texttt{average-few-shot} curve shows the mean accuracy across the three few-shot prompting strategies (\texttt{FS}, \texttt{FS-2}, and \texttt{FS-3}).}
\label{fig:few_shot}
\end{figure*}

\begin{table*}[t]
\centering
\resizebox{\linewidth}{!}{
\begin{tabular}{l|rrrrrr|rrrrrr}
\toprule
\multicolumn{1}{l}{}& \multicolumn{6}{c}{Carry-over Acc.} & \multicolumn{6}{c}{Position Add Acc.} \\
\multicolumn{1}{l}{Task Type} & ZS & S & FS & FS-2 & FS-3 & \multicolumn{1}{r}{E} & ZS & S & FS & FS-2 & FS-3 & E \\
\multicolumn{1}{l}{Models} & & & & & &\multicolumn{1}{r}{} & & & & & & \\
\midrule
Llama3-8b-it & 15.89 & 0.20 & 8.42 & 15.38 & 16.68 & 13.54 & 16.25 & 0.07 & 7.15 & 15.00 & 14.34 & 12.32 \\
Llama3.1-8b-it & 21.96 & 0.25 & 8.84 & 15.33 & 12.46 & 24.80 & 20.38 & 0.10 & 7.92 & 12.91 & 10.14 & 23.61 \\
Llama3.2-11b-it & 17.35 & 0.26 & 9.04 & 19.70 & 13.97 & 27.47 & 16.60 & 0.12 & 8.29 & 18.92 & 12.57 & 27.13 \\
\midrule
Qwen1.5-7b-it & 47.44 & 0.09 & 3.09 & 6.40 & 5.36 & 7.51 & 46.78 & 0.05 & 2.66 & 5.98 & 4.62 & 8.00 \\
Qwen2-7b-it & 62.94 & 0.06 & 28.36 & 57.22 & 32.35 & 70.83 & 60.03 & 0.05 & 23.65 & 48.34 & 28.25 & 68.80 \\
Qwen2.5-7b-it & 74.49 & 0.13 & 38.28 & 55.08 & 41.54 & 72.09 & 71.39 & 0.11 & 33.16 & 48.12 & 36.11 & 71.53 \\
\bottomrule
\end{tabular}
}
\caption{ \textbf{Impact of Different Knowledge Intervention Strategies.} Contrary to expectations, providing explicit rules (few-shot conditions) significantly reduces performance compared to zero-shot baseline, e.g. \texttt{Qwen2.5-7b-it} drop 29.49\%. However, when models explain their reasoning before computation (\texttt{Explain-and-Do}), performance remains comparable to zero-shot levels. \textit{ZS = Zero-Shot}, \textit{FS = Few-Shot}, \textit{E = Explain-and-Do}.}
\label{tab:rule_provision}
\end{table*}

\subsection{RQ2: What factors modulate grasping?}
\label{sec:rq2}

The preceding analysis (RQ1) demonstrated LLMs' significant deficiencies in internalizing elementary addition rules. To further understand the nature of these limitations and explore potential avenues for improvement, RQ2 investigates factors that might modulate LLMs' ability to grasp these rules. We examine two primary categories of interventions: (1) prompt-level strategies, including the provision of explicit rules and the use of self-explanation prompts, and (2) parameter-level modifications through task-specific fine-tuning.

\subsubsection{Explicit Rule Provision}
Building on RQ1's finding that LLMs struggle with genuine arithmetic understanding, this subsection investigates whether explicit rule provision can enhance their performance. We evaluated LLMs under several prompt-level interventions: few-shot prompting with definitions of addition principles and examples of varying digit lengths (denoted \texttt{Few-Shot}, \texttt{Few-Shot-2}, and \texttt{Few-Shot-3}), and an \texttt{Explain-and-Do} strategy, where models first articulate their problem-solving approach. Results are presented in \Cref{tab:rule_provision} and \Cref{fig:few_shot}.

Our investigation reveals a counterintuitive finding: providing LLMs with abstract addition rules consistently degraded performance compared to zero-shot settings. This suggests LLMs favor memorizing token patterns over abstracting principles. When faced with human-articulated rules (e.g., ``carry the 1''), models struggle to operationalize them, defaulting to pre-trained pattern-matching. This preference explains performance disparities between numerical and symbolic tasks and observed commutativity violations. In contrast, the \texttt{Explain-and-Do} strategy—prompting models to first articulate their reasoning—generally maintained performance near zero-shot levels. These findings indicate current LLMs are predominantly optimized for pattern recognition, not abstract rule learning, highlighting a divergence from human mathematical cognition.

Architectural differences among LLM families also influence how models respond to interventions. For instance, Qwen's strong zero-shot performance, paired with its sharp decline when given explicit rules, could indicate a knowledge system highly optimized during pre-training for high-frequency numerical patterns, possibly from extensive exposure to code or tabular data. This optimization would make it efficient at recall but brittle when prompts introduce abstract principles that conflict with these ingrained heuristics. Conversely, Llama's relative success in adapting to rules, especially with the \texttt{Explain-and-Do} prompt, may point to a pre-training or fine-tuning philosophy that fosters more flexible, deliberative reasoning pathways, allowing it to better integrate novel instructions even if its initial pattern recognition is less precise (\Cref{tab:rule_provision}). The importance of the tuning paradigm is further underscored when comparing base versus instruction-tuned models. As detailed in the Appendix, base models performed significantly worse, often failing to follow the prompt format entirely. This reveals that while instruction-tuning is a prerequisite for attempting in-context learning, it does not confer a genuine grasp of the arithmetic rule itself, reinforcing our finding that current methods fail to induce robust procedural understanding.

\begin{table*}[t]
\centering
\resizebox{\linewidth}{!}{
\begin{tabular}{lll|rrr|rrr|rrr|r}
\toprule
&Fine-Tuning Type& \multicolumn{1}{l}{Dataset Domain} & \multicolumn{3}{c}{Overall Acc.} & \multicolumn{3}{c}{Position Add Acc.} & \multicolumn{3}{c}{Carry-over Acc.}& \multicolumn{1}{c}{Map Acc.} \\
Task Type & & & ZS & S & $\Delta$ & ZS & S & $\Delta$ & ZS & S & $\Delta$& S \\
Models & & & & & & & & & & & \\
\midrule
Qwen2.5-7B-it & - & - & 83.00 & 0.58 & -82.41 & 71.39 & 0.11 & -71.28 & 74.49 & 0.13 & -74.37 & 0.57 \\
\midrule
Eurus2-7B-SFT & SFT & Domain Specific & 83.21 & 0.42 & -82.79 & 81.21 & 3.19 & -78.02 & 82.28 & 6.87 & -75.41 &  -\\
Eurus2-7B-PRIME & RL(PRM) & Domain Specific & 94.11 & 1.03 & -93.08 & 91.59 & 3.10 & -88.49 & 92.51 & 3.11 & -89.40 & - \\
DS-R1-Distill-Qwen-7B & RL(Reasoning) & General & 74.76 & 6.88 & -67.88 & 65.38 & 33.41 & -31.97 & 64.27 & 31.52 & -32.75 &-  \\
\midrule
Qwen2.5-7B-it & SFT & Task Specific (Numerical) & 97.17 & 0.00 & -97.17 & 87.91 & 0.25 & -87.66 & 89.51 & 1.26 & -88.25 & 8.21 \\
Qwen2.5-7B-it & RL(DPO) & Task Specific (Numerical) & 95.32 & 0.37 & -94.95 & 86.23 & 1.17 & -85.06 & 87.75 & 2.35 & -85.40 & 2.25 \\
Qwen2.5-7B-it & RL(SFT+DPO) & Task Specific (Numerical) & 96.95 & 0.28 & -96.67 & 84.48 & 0.29 & -84.19 & 85.52 & 0.61 & -84.91 & 0.10 \\
Qwen2.5-7B-it & SFT & Task Specific (Symbolic) & 0.00 & 30.66 & +30.66 & 3.40 & 3.89 & +0.49 & 6.71 & 6.98 & +0.27 & 23.49 \\
Qwen2.5-7B-it & RL(DPO) & Task Specific (Symbolic) & 50.73 & 24.10 & -26.63 & 47.71 & 3.48 & -44.23 & 48.40 & 6.37 & -42.03 & 19.84 \\
Qwen2.5-7B-it & RL(SFT+DPO) & Task Specific (Symbolic) & 12.32 & 2.85 & -9.47 & 9.31 & 0.58 & -8.73 & 9.70 & 1.13 & -8.57 & 2.00 \\
\bottomrule
\end{tabular}
}
\caption{ \textbf{Impact of Fine-Tuning Approaches on Arithmetic Capabilities.} Different fine-tuning strategies and dataset domains yield distinct trade-offs between performance and generalization. While SFT achieves highest numerical accuracy, it shows minimal transfer to symbolic tasks. RL-based approaches demonstrate better generalization but lower absolute performance. Task-specific training on numerical data excels within-domain but fails to transfer, whereas general-domain training (e.g., DS-R1-Distill) enables broader generalization through its diverse training objectives, suggesting the importance of training paradigm design in developing robust mathematical capabilities.}
\label{tab:finetuning}
\end{table*}

\subsubsection{Rule Internalization}
We then investigated if parameter-level modifications via fine-tuning could improve LLMs' internalization of arithmetic rules, moving beyond mere pattern matching. We explored various fine-tuning strategies: SFT, RL with Direct Preference Optimization (DPO) \citep{rafailov2023direct}, and a hybrid RPO (SFT+DPO) \citep{pang2024iterativce}. Model performance was assessed on both numerical and symbolic addition post-fine-tuning, and benchmarked against specialized mathematical reasoning models like \texttt{Eurus2} \citep{cui2025process}, \texttt{OpenAI o1} \citep{openai2024o1}, \texttt{DeepSeek R1} \citep{deepseekai2025deepseekr1}, and their distilled counterparts. For RL, training data comprised model responses, with correct and incorrect answers from our dataset serving as positive and negative examples, respectively (details in \Cref{app:ft-conf}).

Fine-tuning experiments (\Cref{tab:finetuning}) revealed clear trade-offs. Task-specific SFT boosted performance on in-domain numerical addition but failed to generalize to symbolic one, indicating that SFT primarily reinforces pattern matching tied to data. Conversely, RL-based methods (DPO and RPO) achieved better generalization to symbolic inputs, albeit with lower absolute accuracy on the fine-tuned numerical task. Notably, the RPO still struggled with symbolic transfer, suggesting SFT's propensity for pattern matching can overshadow RL's generalization benefits. These findings imply that standard fine-tuning, particularly SFT, optimizes for surface-level pattern recognition over the abstraction of underlying arithmetic principles.

Supporting this, models fine-tuned on general-domain reasoning objectives (e.g., DS-R1-Distill) demonstrated more robust generalization to symbolic tasks. This improved transfer is likely due to training objectives that promote extended reasoning, highlighting the training paradigm's crucial role in fostering generalizable mathematical skills. In contrast, domain-specific models like \texttt{Eurus2-SFT} and \texttt{Eurus2-PRIME}, despite excelling at complex numerical tasks within their domain, showed limited transfer to symbolic addition. However, \texttt{Eurus2-PRIME} generalized better than \texttt{Eurus2-SFT}. This suggests RL-based signals can aid in abstracting principles, though balancing specialization with generalization remains challenging.

Specialized reasoning models (\Cref{tab:internlization}) offered further insights. These models typically showed less performance degradation on symbolic addition compared to standard LLMs, suggesting that training on prolonged or complex reasoning tasks can foster better abstraction of arithmetic principles. Yet, this improved abstraction may entail a trade-off: some reasoning-focused architectures sacrificed accuracy on elementary computations (\Cref{fig:zero_shot_symbolic_digit_count}), potentially by ``over-thinking'' simple problems despite excelling at complex ones. This pattern underscores how architectural design and training objectives critically shape the balance between foundational computational skills and higher-order reasoning.

\begin{table}[t]
\centering
\resizebox{\linewidth}{!}{
\begin{tabular}{lrrrrrr}
\toprule
& \multicolumn{3}{c}{Position Add Acc.} & \multicolumn{3}{c}{Carry-over Acc.} \\
Task Type & ZS & S & $\Delta$ & ZS & S & $\Delta$ \\
\midrule
Gemini2.0-pro-exp & 69.52 & 4.19 & -65.33 & 77.36 & 7.07 & -70.29 \\
Gemini2.5-pro-exp (thinking) & 88.97 & 19.80 & -69.17 & 88.49 & 24.56 & -63.93 \\
\midrule
Llama3.3-70b-it & 73.82 & 0.77 & -73.05 & 75.00 & 2.43 & -72.57 \\
DS-R1-Distill-Llama-70B & 68.91 & 42.94 & -25.97 & 68.56 & 40.75 & -27.81 \\
\midrule
Llama3.1-8b-it & 20.38 & 0.10 & -20.27 & 21.96 & 0.25 & -21.72 \\
DS-R1-Distill-Llama-8B & 45.54 & 39.55 & -5.99 & 44.16 & 35.09 & -9.07 \\
\bottomrule
\end{tabular}
}
\caption{ \textbf{LLMs' Understanding of Addition Principles.} Models achieve high accuracy(\%) on standard numerical tasks (zero-shot) but show severe degradation when tested on symbolic representations, both for carry operations and digit addition. This stark contrast suggests that models only grasp principles in numerical form and fail to generalize to abstract representations.}
\label{tab:internlization}
\end{table}

\section{Conclusion}
Our empirical results of two-integer addition task reveal that \textbf{LLMs fail to grasp elementary addition rules, still relying instead on surface-level pattern matching.} This conclusion is evidenced by: (1) a collapse in accuracy (e.g., from $\ge99\%$ to $\le7.5\%$) when standard digits are replaced with novel symbols, demonstrating a lack of notation invariance; (2) non-monotonic accuracy scaling with operand length, suggesting specific memorization over consistent carry-propagation; and (3) systematic commutativity violations, which contradict genuine rule grasp. These findings collectively indicate that LLMs' success on complex math benchmarks may mask a superficial understanding of foundational rules, and our preliminary results show this failure generalizes to other arithmetic operations.

Interventions further highlight these deficits: providing formal rules from human knowledge paradoxically degrades performance (by up to 81.2\%), while prompting models to \texttt{Explain-and-Do} merely preserves baseline scores. Task-specific SFT boosts numeric accuracy but fails to generalize to symbolic tasks; conversely, RL shows better symbolic transfer but at the cost of lower absolute accuracy. This suggests a fundamental misalignment between human-like abstract rule learning and the pattern-matching heuristics LLMs develop during pre-training.

The implications are significant: current benchmarks, rewarding final answers over rule fidelity, risk inflating perceived LLM competence. Future evaluations must test notation invariance, scaling consistency, and algebraic integrity. Model design should explore explicit symbolic manipulation or execution-grounded reasoning. Bridging the pattern recall-rule abstraction gap is crucial for genuine mathematical understanding in LLMs.

\section*{Acknowledgements}
We would like to express our gratitude to the Gemini Developer API Team at Google for providing the extensive Gemini API access, which greatly facilitated our research. We also thank the anonymous reviewers for their insightful feedback. This work was funded by the Scientific Research Project of Westlake University (Grant No. WU2024B003).

\section*{Ethical Considerations}

While our research primarily focuses on the mathematical reasoning capabilities of LLMs, which not directly involve ethical considerations. However, the implications of our findings extend to broader ethical concerns in AI deployment. We highlight following key areas:

\paragraph{Why arithmetic robustness matters.}
Elementary addition underpins many downstream computations. A model that answers graduate-level problems yet violates commutativity can silently corrupt applications that rely on implicit arithmetic, including dose calculation, portfolio rebalancing, and automated bidding. This gap between perceived and actual competence creates a direct safety hazard.

\paragraph{Inflated competence metrics.}
Public leaderboards optimise for final-answer accuracy rather than rule fidelity. Our results show that such metrics can conceal thousands of systematic arithmetic errors. Deploying models on the basis of these scores may therefore foster unwarranted confidence and expose users to financial or physical harm.

\paragraph{Recommendations for high-stakes deployment.}
Before adoption in safety-critical settings, developers should (i) report notation-invariance and algebraic-integrity scores alongside aggregate benchmarks, (ii) document failure modes such as the symbol-mapping collapse identified here, and (iii) install run-time monitors that flag out-of-distribution numeric inputs. These measures align claimed capability with real-world reliability.

\paragraph{Toward stronger evaluation standards.}
The field needs public, reproducible suites that test formal properties directly, not just end-to-end accuracy. Without such standards, the gap between apparent and actual mathematical competence will widen and public trust in AI will erode.

With these considerations in mind, we would highlight the importance and significance these findings have for the future of AI systems. As LLMs are increasingly integrated into various domains, ensuring their reliability and robustness in fundamental tasks like arithmetic is crucial for safe and effective deployment.

\section*{Limitations}

\paragraph{Scope of mathematical operations.}
Our study targets two-integer addition because it offers a clean probe of rule learning. Preliminary experiments suggest similar failures in subtraction, multiplication, and symbolic logic, but verifying those trends remains future work.

\paragraph{Range of intervention techniques.}
We evaluate prompt engineering, SFT, and preference-based RL. Alternative strategies—such as modular arithmetic heads, execution-augmented decoding, or neuro-symbolic hybrids—may yield different generalisation patterns that we have not explored.

\paragraph{External validity of the synthetic dataset.}
The symbol-mapping protocol strips away contextual cues that may aid reasoning. In real documents, numeric reasoning is embedded in richer text, so model behaviour could differ. Future studies should embed the same invariance checks in realistic narratives such as medical charts or financial statements.

\paragraph{Sampling constraints.}
API costs limited us to fewer than ten stochastic decodes for some proprietary models. Although the observed failure margins are large, denser sampling would narrow confidence intervals.

\paragraph{Mechanistic understanding.}
We observe strong evidence of pattern matching rather than rule induction, yet the circuit-level mechanisms remain unidentified. Tracing these mechanisms and designing architectures that promote rule abstraction are important directions for future research.

\FloatBarrier
\bibliography{custom}

\appendix
\section{Appendix}

\subsection{AI Use Statement}
This research utilized AI assistance for code debugging and grammatical refinement. All experimental designs, analyses, results, and conclusions were developed independently by the authors without generative AI input. We employed AI tools solely for technical implementation support and language polishing to ensure clear communication of our findings.

\subsection{Experimental Setup}
\label{app:experimental_setup}

Our evaluation framework utilized the SGLang platform through the official Docker container \texttt{lmsysorg/sglang}~\citep{zheng2024sglang}. For statistical robustness, most models underwent 10 repeated evaluations per test example using a temperature setting of 0.7 across the full dataset. Due to computational and budget constraints, select models including \texttt{GPT4-o}, \texttt{Claude-3.5-Sonnet}, \texttt{QwQ-32B-Preview}, \texttt{Deepseek-R1} and its variants were evaluated once on the test split only.

For assessing \texttt{Position Addition} and \texttt{Carry-over} accuracy, we used \texttt{Phi-4}~\citep{abdin2024phi4} as an independent generative evaluator following~\citet{zhang2024generative}. Solutions were evaluated by feeding them to the evaluator to determine carry-over and position addition correctness, using the first token as the prediction.

We conducted comprehensive evaluations across all model variants in both zero-shot and symbolic settings, with complete results presented in \Cref{tab:zero_shot_symbolic_full}.

\subsection{Clarification on the Digit-Scaling Consistency Diagnostic}
\label{app:scaling_clarification}
Our digit-scaling consistency diagnostic posits that accuracy for a rule-based system should be non-increasing with operand length. This premise rests on the observation that LLM performance does not align with either of two coherent models of behavior. The first is a ``stable calculator'' model, where a perfectly internalized algorithm would yield constant, high accuracy irrespective of operand length. The second is a ``human-aligned'' model, where cognitive load increases with complexity, leading to a monotonic decrease in performance. Our key finding—the non-monotonic 'drop-rebound' pattern, where accuracy falls for 4–6 digit numbers but then improves for 8–10 digits (\Cref{fig:zero_shot_symbolic_digit_count})—is inconsistent with \emph{both} models. This erratic scaling strongly suggests that models are not applying a single, coherent rule but are instead relying on a patchwork of length-specific heuristics and memorized patterns.

\subsection{Extended Experiments for Generalizability}
\label{app:generalizability_experiments}
To address concerns about the artificiality of our primary task and to test the generalizability of our findings, we conducted an experiment embedding our addition problems into natural language templates, styled after the GSM8K-Symbolic benchmark~\citep{mirzadeh2025gsmsymbolic}. This `semantic' setting provides a more naturalistic context. As shown in \Cref{tab:semantic_results}, a \texttt{Qwen2.5-7B-it} model demonstrates consistently higher accuracy in the semantic setting. However, the fundamental failure mode persists: the accuracy curve still exhibits the non-monotonic `drop-rebound' pattern, confirming that our core findings are not an artifact of a single synthetic task.

\begin{table}[h]
\centering
\caption{ \textbf{Per-digit accuracy of \texttt{Qwen2.5-7B-it} on numerical addition tasks in a standard (zero-shot) vs. natural language (`semantic') context.} While performance is higher in the semantic context, the non-monotonic scaling pattern persists, indicating a continued reliance on heuristics over a general rule.}
\label{tab:semantic_results}
\resizebox{0.5\linewidth}{!}{
\begin{tabular}{rrr}
\toprule
\textbf{Digits} & \textbf{zero-shot} & \textbf{semantic} \\
\midrule
1 & 99.45 & 100.00 \\
2 & 99.60 & 99.99 \\
3 & 98.95 & 99.70 \\
4 & 95.11 & 98.72 \\
5 & 91.36 & 98.00 \\
6 & 89.29 & 97.00 \\
7 & 86.53 & 95.85 \\
8 & 83.74 & 95.09 \\
9 & 79.18 & 92.99 \\
10 & 79.26 & 86.56 \\
11 & 78.84 & 82.54 \\
12 & 78.83 & 81.76 \\
13 & 79.03 & 82.10 \\
14 & 75.76 & 83.43 \\
15 & 71.32 & 79.64 \\
16 & 67.90 & 80.44 \\
18 & 60.00 & 60.00 \\
19 & 51.50 & 56.48 \\
20 & 57.73 & 68.01 \\
\bottomrule
\end{tabular}}
\end{table}

\subsection{Probing ICL and Learning Dynamics}
\label{app:probing_dynamics}
To deeper understand the failure modes of in-context learning (ICL) and rule application, we conducted three targeted experiments.

\paragraph{Base vs. Instruction-Tuned Models.} We compared the ICL performance of a base model (\texttt{Qwen2.5-7B-Base}) with its instruction-tuned counterpart (\texttt{Qwen2.5-7B-it}). As shown in \Cref{tab:icl_base_instruct}, the base model struggles to follow the task format, highlighting that instruction-tuning is a prerequisite for even attempting ICL. However, even the instruction-tuned model's performance degrades when provided with explicit rules (see \Cref{tab:rule_provision}), reinforcing that instruction-tuning does not confer true rule-based generalization.

\paragraph{Alternative Mapping Schemes.} We tested performance with a ``shift-cipher'' mapping (e.g., `0' $\leftrightarrow$ `1'), which creates stronger conflicts with ingrained numerical patterns. As shown in \Cref{tab:shift_cipher}, performance on this task collapses, confirming that models rely on familiar token patterns rather than abstract, symbol-agnostic rules.

\paragraph{Explicit Intermediate Reasoning.} We prompted models to output the ``sum up to now'' at each computational step to make the iterative process explicit. This intervention did not improve performance (\Cref{tab:sum_up}), suggesting the failure is fundamental to the rule-application process itself, not merely an issue of tracking intermediate state.

\begin{table}[h]
\centering
\caption{ \textbf{ICL performance for Base vs. Instruction-Tuned models (\texttt{Qwen2.5-7B}).} Results show the base model struggles to perform the task, while fine-tuning on symbolic data (\texttt{sft+S}) enables symbolic reasoning but at the cost of zero-shot numeric performance.}
\label{tab:icl_base_instruct}
\resizebox{\columnwidth}{!}{
\begin{tabular}{lrrrrrrr}
\toprule
\textbf{Digits} & \multicolumn{1}{c}{\shortstack[c]{ZS \\ Base}} & \multicolumn{1}{c}{\shortstack[c]{ZS \\ IT}} & \multicolumn{1}{c}{\shortstack[c]{ZS \\ SFT-ZS}} & \multicolumn{1}{c}{\shortstack[c]{S \\ Base}} & \multicolumn{1}{c}{\shortstack[c]{S \\ IT}} & \multicolumn{1}{c}{\shortstack[c]{S \\ SFT-S}} & \multicolumn{1}{c}{\shortstack[c]{S \\ SFT-ZS}} \\
\midrule
1-3 & 94.61 & 99.33 & 99.48 & 40.45 & 29.11 & 96.85 & 0.00 \\
4-6 & 85.27 & 91.92 & 99.39 & 0.10 & 0.07 & 69.41 & 0.00 \\
7-9 & 79.51 & 83.15 & 97.78 & 0.00 & 0.00 & 23.13 & 0.00 \\
10-12 & 72.48 & 78.98 & 96.55 & 0.00 & 0.00 & 1.21 & 0.00 \\
13-16 & 64.48 & 71.25 & 96.71 & 0.00 & 0.00 & 0.00 & 0.00 \\
18-20 & 53.39 & 56.41 & 86.78 & 0.00 & 0.00 & 0.00 & 0.00 \\
\bottomrule
\end{tabular}}
\end{table}

\begin{table}[h]
\centering
\caption{ \textbf{Performance of \texttt{Qwen2.5-7B-it} with a ``shift-cipher'' mapping.} The near-total collapse in accuracy compared to the standard symbolic task highlights the model's reliance on familiar token patterns.}
\label{tab:shift_cipher}
\resizebox{0.7\linewidth}{!}{
\begin{tabular}{rrrr}
\toprule
\textbf{Digits} & \textbf{symbolic} & \textbf{shift-cipher} & \textbf{zero-shot} \\
\midrule
1-3 & 55.43 & 1.30 & 85.13 \\
4-6 & 24.33 & 0.00 & 78.97 \\
7-9 & 9.37 & 0.00 & 73.63 \\
10-12 & 3.73 & 0.00 & 72.30 \\
13-16 & 1.87 & 0.00 & 68.43 \\
18-20 & 0.43 & 0.00 & 60.17 \\
\bottomrule
\end{tabular}}
\end{table}

\begin{table}[h]
\centering
\caption{ \textbf{Performance of \texttt{Qwen2.5-7B-it} with an explicit ``sum-up'' intermediate reasoning step.} The lack of improvement indicates the core reasoning failure is not simply due to memory or state-tracking limitations.}
\label{tab:sum_up}
\resizebox{0.7\linewidth}{!}{
\begin{tabular}{rrrr}
\toprule
\textbf{Digits} & \textbf{symbolic} & \textbf{symbolic-sumup} & \textbf{zero-shot} \\
\midrule
1-3 & 55.43 & 29.30 & 85.13 \\
4-6 & 24.33 & 0.07 & 78.97 \\
7-9 & 9.37 & 0.00 & 73.63 \\
10-12 & 3.73 & 0.00 & 72.30 \\
13-16 & 1.87 & 0.00 & 68.43 \\
18-20 & 0.43 & 0.00 & 60.17 \\
\bottomrule
\end{tabular}}
\end{table}

\subsection{Fine-Tuning Configuration}
\label{app:ft-conf}

Our investigation employed three fine-tuning approaches: standard DPO, RPO (combining DPO with SFT), and pure SFT. Each approach shared core configuration elements while varying key method-specific parameters.

\paragraph{Base Configuration.} The base configuration utilized a batch size of 1 sample per device with 4 gradient accumulation steps (effective batch size of 4). Training ran for 1 epoch using cosine learning rate scheduling with 10\% warmup steps. We implemented BF16 mixed precision and non-reentrant gradient checkpointing, evaluating on a 1\% validation set every 500 steps. Flash Attention 2 optimized computation efficiency.

\paragraph{Distributed Training.} Training leveraged DeepSpeed ZeRO-3 with 8 processes per machine. The implementation included CPU optimizer state offloading, gradient clipping at 1.0, 16-bit parameter saving, and static process coordination through DeepSpeed's rendezvous mechanism.

\paragraph{Method-specific Parameters.}
\begin{itemize}[leftmargin=*, itemindent=0em, itemsep=0pt, parsep=0pt, topsep=0pt, partopsep=0pt]
\item Standard DPO: Learning rate $5.0\times10^{-6}$, $\beta=0.0$, sigmoid loss function
\item RPO: DPO settings with $\beta=1.0$ for integrated preference modeling and SFT
\item SFT: Learning rate $1.0\times10^{-4}$ for supervised training
\end{itemize}

All approaches utilized full-parameter fine-tuning through DeepSpeed ZeRO-3. For preference learning (DPO/RPO), we initialized reference models from SFT checkpoints with preference loss weight ($\lambda_{\text{ftx}}$) set to 1.0.

\paragraph{Infrastructure.} Training infrastructure consisted of 4 NVIDIA A100 GPUs (80GB each), with complete fine-tuning requiring approximately 15 hours per run.

\begin{figure*}[ht]
\scriptsize
\begin{tcolorbox}[width=1\linewidth,title={\textbf{Prompt Template for Zero-Shot Setting}}]
\textbf{Context:} \\
You are a helpful AI assistant. \\

\textbf{Instruction:} \\
Present your solution in the following format: \\
1. Let's think step by step. \\
2. Final Answer: Express using LaTeX notation \texttt{\textbackslash boxed\{answer\}} \\

\textbf{Question:} \\
\texttt{\%s + \%s = \textbackslash boxed\{?\}}
\end{tcolorbox}
\caption{ \textbf{Zero-Shot Setting Prompt Template.} Example prompt template for zero-shot addition tasks, providing context, instructions, and question format for LLMs.}
\label{fig:zero_shot_template}
\end{figure*}

\begin{figure*}[ht]
\scriptsize
\begin{tcolorbox}[width=1\linewidth,title={\textbf{Prompt Template for Few-Shot Setting}}]
\textbf{Context:} \\
You are a helpful AI assistant. \\

\textbf{Instruction:} \\
Present your solution in the following format: \\
1. First, compute the sum of the two numbers, working from right to left using place values. \\
2. Then, for each place value, add the digits in the same place value column, and carry over if the sum is greater than 9. \\
3. Iterate this process from right to left until all place values are added. \\
4. Final Answer: Express using LaTeX notation \texttt{\textbackslash boxed\{answer\}}. \\

\textbf{Examples:} \\
1. Compute \texttt{1996 + 126 = \textbackslash boxed\{?\}} \\
\texttt{Let's solve 1996 + 126 step by step, working from right to left using place values.} \\
\begin{itemize}[leftmargin=*, itemindent=0em, itemsep=0pt, parsep=0pt, topsep=0pt, partopsep=0pt]
\item For the ones place: 6 + 6 = 12. Write down 2 in the ones place and carry over 1 to the tens place.
\item For the tens place: 9 + 2 + 1 = 12. Write down 2 in the tens place and carry over 1 to the hundreds place.
\item For the hundreds place: 9 + 1 + 1 = 11. Write down 1 in the hundreds place and carry over 1 to the thousands place.
\item For the thousands place: 1 + 1 = 2.
\item Putting it all together: \texttt{2 * 1000 + 1 * 100 + 2 * 10 + 2 * 1 = 2000 + 100 + 20 + 2 = 2122}.
\end{itemize}
Therefore, \texttt{1996 + 126 = \textbackslash boxed\{2122\}}. \\

2. Compute \texttt{1994 + 222 = \textbackslash boxed\{?\}} \\
\texttt{Let's solve 1994 + 222 step by step, working from right to left using place values.} \\
\begin{itemize}[leftmargin=*, itemindent=0em, itemsep=0pt, parsep=0pt, topsep=0pt, partopsep=0pt]
\item For the ones place: 2 + 4 = 6.
\item For the tens place: 2 + 9 = 11. Write down 1 in the tens place and carry over 1 to the hundreds place.
\item For the hundreds place: 2 + 9 + 1 = 12. Write down 2 in the hundreds place and carry over 1 to the thousands place.
\item For the thousands place: 1 + 1 = 2.
\item Putting it all together: \texttt{2 * 1000 + 2 * 100 + 1 * 10 + 6 * 1 = 2000 + 200 + 10 + 6 = 2216}.
\end{itemize}
Therefore, \texttt{1994 + 222 = \textbackslash boxed\{2216\}}. \\

\textbf{Question:} \\
\texttt{\%s + \%s = \textbackslash boxed\{?\}}
\end{tcolorbox}
\caption{ \textbf{Few-Shot Setting Prompt Template.} Example prompt template for few-shot addition tasks, providing context, instructions, examples, and question format for LLMs.}
\label{fig:few_shot_template}
\end{figure*}

\begin{figure*}[ht]
\scriptsize
\begin{tcolorbox}[width=1\linewidth,title={\textbf{Prompt Template for Explain-and-Do Setting}}]
\textbf{Context:} \\
You are a helpful AI assistant. \\

\textbf{Instruction:} \\
Present your solution in the following format: \\
1. First, comprehensively explain how to do addition with both positive integers. \\
2. Then, let's analyze the problem step by step following your explanation. \\
3. Final Answer: Express using LaTeX notation \texttt{\textbackslash boxed\{answer\}}. \\

\textbf{Question:} \\
\texttt{\%s + \%s = \textbackslash boxed\{?\}}
\end{tcolorbox}
\caption{ \textbf{Explain-and-Do Setting Prompt Template.} Example prompt template for explain-and-do addition tasks, providing context, instructions, and question format for LLMs.}
\label{fig:explain_and_do_template}
\end{figure*}

\begin{figure*}[ht]
\scriptsize
\begin{tcolorbox}[width=1\linewidth,title={\textbf{Prompt Template for Symbolic Setting}}]
\textbf{Context:} \\
You are a helpful AI assistant. Your task is to perform addition within a custom symbolic system in a simple and clear manner. \\

\textbf{Symbolic System Definition:} \\
This system comprises ten symbols: \{u, d, a, i, h, v, e, y, r, c\}. The addition operation (+) between these symbols is defined as follows: \\

\texttt{u + u = u} \\
\texttt{d + u = d \hspace{1em} d + d = a} \\
\texttt{a + u = a \hspace{1em} a + d = i \hspace{1em} a + a = h} \\
\texttt{i + u = i \hspace{1em} i + d = h \hspace{1em} i + a = v \hspace{1em} i + i = e} \\
\texttt{h + u = h \hspace{1em} h + d = v \hspace{1em} h + a = e \hspace{1em} h + i = y \hspace{1em} h + h = r} \\
\texttt{v + u = v \hspace{1em} v + d = e \hspace{1em} v + a = y \hspace{1em} v + i = r \hspace{1em} v + h = c \hspace{1em} v + v = du} \\
\texttt{e + u = e \hspace{1em} e + d = y \hspace{1em} e + a = r \hspace{1em} e + i = c \hspace{1em} e + h = du \hspace{1em} e + v = dd \hspace{1em} e + e = da} \\
\texttt{y + u = y \hspace{1em} y + d = r \hspace{1em} y + a = c \hspace{1em} y + i = du \hspace{1em} y + h = dd \hspace{1em} y + v = da \hspace{1em} y + e = di \hspace{1em} y + y = dh} \\
\texttt{r + u = r \hspace{1em} r + d = c \hspace{1em} r + a = du \hspace{1em} r + i = dd \hspace{1em} r + h = da \hspace{1em} r + v = di \hspace{1em} r + e = dh \hspace{1em} r + y = dv \hspace{1em} r + r = de} \\
\texttt{c + u = c \hspace{1em} c + d = du \hspace{1em} c + a = dd \hspace{1em} c + i = da \hspace{1em} c + h = di \hspace{1em} c + v = dh \hspace{1em} c + e = dv \hspace{1em} c + y = de \hspace{1em} c + r = dy \hspace{1em} c + c = dr} \\

\textbf{Instruction:} \\
Present your solution in the following format: \\
1. \textbf{Align:} Arrange the two input strings vertically, aligning their rightmost symbols. \\
2. \textbf{Columnar Addition:} Starting from the rightmost column (least significant symbols), perform symbol addition using the provided definition. \\
3. \textbf{Carry-over:} If the result of a column's addition is a two-symbol sequence (e.g., 'da'), write down the second symbol (least significant) and carry over the first symbol to the next column on the left. \\
4. \textbf{Iteration:} Repeat steps 2 and 3, moving leftward column by column until all symbols have been added. \\
5. \textbf{Reasoning:} Keep your whole reasoning clear and simple. \\
6. \textbf{Output Format:} Write the final result in the \texttt{\textbackslash boxed\{?\}} placeholder. \\

\textbf{Examples:} \\
1. Compute \texttt{dcce + dae = \textbackslash boxed\{?\}} \\

\textbf{Solution:} \\
\texttt{1. Columnar Addition (right to left):} \\
\texttt{- e + e = da (Write 'a', Carry 'd')} \\
\texttt{- c + a + d = da (Write 'a', Carry 'd')} \\
\texttt{- c + d + d = dd (Write 'd', Carry 'd')} \\
\texttt{- d + d = a} \\
\texttt{2. Result: adaa} \\
\texttt{3. Formatted Output: \textbackslash boxed\{adaa\}} \\

2. Compute \texttt{dcch + aaa = \textbackslash boxed\{?\}} \\

\textbf{Solution:} \\
\texttt{1. Columnar Addition (right to left):} \\
\texttt{- h + a = e} \\
\texttt{- c + a = dd (Write 'd', Carry 'd')} \\
\texttt{- c + a + d = da (Write 'a', Carry 'd')} \\
\texttt{- d + d = a} \\
\texttt{2. Result: aade} \\
\texttt{3. Formatted Output: \textbackslash boxed\{aade\}} \\

\textbf{Your Task:} \\
\texttt{Compute \%s + \%s = \textbackslash boxed\{?\}} \\
\end{tcolorbox}
\caption{ \textbf{Symbolic Setting Prompt Template.} Example prompt template for symbolic addition tasks, providing context, symbolic system definition, instructions, examples, and question format for LLMs.}
\label{fig:symbolic_template}
\end{figure*}

\begin{table*}[t]
\centering
\resizebox{0.9\linewidth}{!}{
\begin{tabular}{lrrrrrrrrr}
\toprule
& \multicolumn{3}{c}{Overall Acc.} & \multicolumn{3}{c}{Position Add Acc.} & \multicolumn{3}{c}{Carry-over Acc.} \\
Task Type & ZS & S & $\Delta$ & ZS & S & $\Delta$ & ZS & S & $\Delta$ \\
Models & & & & & & & & & \\
\midrule
Gemini2.0-pro-exp & 94.88 & 14.21 & -80.67 & 69.52 & 4.19 & -65.33 & 77.36 & 7.07 & -70.29 \\
Gemini2.5-pro-exp (thinking) & 99.16 & 55.99 & -43.17 & 88.97 & 19.80 & -69.17 & 88.49 & 24.56 & -63.93 \\
Gemini2.0-flash-exp & 98.10 & 9.25 & -88.85 & 73.83 & 1.21 & -72.62 & 79.52 & 3.28 & -76.24 \\
Gemini2.0-flash-exp (thinking) & 91.07 & 10.81 & -80.26 & 86.09 & 2.89 & -83.20 & 88.30 & 9.03 & -79.27 \\
\midrule
Claude-3.5-Sonnet & 99.81 & 7.51 & -92.30 & 81.78 & 3.19 & -78.59 & 90.28 & 6.92 & -83.36 \\
GPT-4o & 93.39 & 9.59 & -83.80 & 76.12 & 3.79 & -72.33 & 79.55 & 6.73 & -72.82 \\
O1-preview & 74.28 & - & - & 74.71 & - & - & 74.23 & - & - \\
ERNIE-Speed-8K & 73.78 & 0.29 & -73.49 & 67.66 & 0.07 & -67.59 & 70.89 & 0.21 & -70.68 \\
\midrule
DeepSeek-V2.5 & 95.75 & - & - & 83.78 & - & - & 88.19 & - & - \\
DeepSeek-V3 & 98.92 & 16.14 & -82.78 & 78.55 & 11.98 & -66.57 & 81.14 & 15.23 & -65.91 \\
DeepSeek-R1 & 97.39 & - & - & 70.99 & - & - & 80.58 & - & - \\
\cline{2-10}\\[-5pt]
DeepSeek-R1-Distill-Llama-70B & 74.19 & 27.19 & -47.00 & 68.91 & 42.94 & -25.97 & 68.56 & 40.75 & -27.81 \\
DeepSeek-R1-Distill-Llama-8B & 53.23 & 10.97 & -42.26 & 45.54 & 39.55 & -5.99 & 44.16 & 35.09 & -9.07 \\
DeepSeek-R1-Distill-Qwen-1.5B & 58.16 & 0.66 & -57.50 & 47.85 & 26.16 & -21.69 & 47.16 & 20.79 & -26.37 \\
DeepSeek-R1-Distill-Qwen-7B & 74.76 & 6.88 & -67.88 & 65.38 & 33.41 & -31.97 & 64.27 & 31.52 & -32.75 \\
\midrule
Gemma2-2b-it & 33.41 & - & - & 29.97 & - & - & 30.59 & - & - \\
Gemma2-9b-it & 66.34 & 1.45 & -64.89 & 58.52 & 0.34 & -58.18 & 60.44 & 0.44 & -59.99 \\
Gemma2-27b-it & 83.65 & 2.62 & -81.03 & 74.77 & 0.91 & -73.85 & 76.68 & 0.91 & -75.77 \\
\midrule
Llama2-7b-it & 19.59 & 0.00 & -19.59 & 20.44 & 0.01 & -20.43 & 22.58 & 0.01 & -22.57 \\
\cline{2-10}\\[-5pt]
Llama3-8B-it & 32.95 & 0.24 & -32.70 & 16.25 & 0.07 & -16.18 & 15.89 & 0.20 & -15.69 \\
Llama3-70B-it & 69.15 & 1.62 & -67.53 & 59.84 & 0.39 & -59.45 & 60.22 & 0.70 & -59.52 \\
\cline{2-10}\\[-5pt]
Llama3.1-8B-it & 43.34 & 0.57 & -42.76 & 20.38 & 0.10 & -20.27 & 21.96 & 0.25 & -21.72 \\
Llama3.1-70B-it & 72.58 & 2.51 & -70.07 & 60.13 & 0.52 & -59.61 & 61.05 & 1.33 & -59.71 \\
\cline{2-10}\\[-5pt]
Llama3.2-11B-it & 35.13 & 0.53 & -34.61 & 16.60 & 0.12 & -16.48 & 17.35 & 0.26 & -17.09 \\
\cline{2-10}\\[-5pt]
Llama3.3-70B-it & 79.63 & 4.01 & -75.61 & 73.82 & 0.77 & -73.05 & 75.00 & 2.43 & -72.57 \\
\midrule
Qwen1.5-7B-Chat & 56.31 & 0.18 & -56.14 & 46.78 & 0.05 & -46.73 & 47.44 & 0.09 & -47.34 \\
Qwen1.5-72B-Chat & 34.29 & 0.53 & -33.75 & 62.28 & 0.09 & -62.20 & 67.28 & 0.14 & -67.13 \\
\cline{2-10}\\[-5pt]
Qwen2-7B-it & 72.50 & 0.24 & -72.26 & 60.03 & 0.05 & -59.98 & 62.94 & 0.06 & -62.88 \\
Qwen2-72B-it & 59.06 & 2.50 & -56.56 & 82.82 & 0.21 & -82.62 & 86.62 & 0.26 & -86.36 \\
\cline{2-10}\\[-5pt]
Qwen2.5-1.5B-it & 47.75 & - & - & 32.54 & - & - & 33.67 & - & - \\
Qwen2.5-3B-it & 70.27 & - & - & 54.49 & - & - & 57.98 & - & - \\
Qwen2.5-7B-it & 83.00 & 0.58 & -82.41 & 71.39 & 0.11 & -71.28 & 74.49 & 0.13 & -74.37 \\
Qwen2.5-14B-it & 87.45 & - & - & 77.56 & - & - & 80.36 & - & - \\
Qwen2.5-32B-it & 95.15 & - & - & 90.41 & - & - & 91.28 & - & - \\
Qwen2.5-72B-it & 96.07 & 5.97 & -90.10 & 88.19 & 2.09 & -86.10 & 89.78 & 4.12 & -85.67 \\
\cline{2-10}\\[-5pt]
QwQ-32B-Preview & 70.59 & 11.12 & -59.47 & 71.68 & 19.09 & -52.59 & 73.22 & 20.71 & -52.51 \\
\cline{2-10}\\[-5pt]
Eurus2-7B-SFT & 83.21 & 0.42 & -82.79 & 81.21 & 3.19 & -78.02 & 82.28 & 6.87 & -75.41 \\
Eurus2-7B-PRIME & 94.11 & 1.03 & -93.08 & 91.59 & 3.10 & -88.49 & 92.51 & 3.11 & -89.40 \\
\cline{2-10}\\[-5pt]
qwen2.5-7b-dpo-sft-S & 12.32 & 2.85 & -9.47 & 9.31 & 0.58 & -8.73 & 9.70 & 1.13 & -8.57 \\
qwen2.5-7b-dpo-sft-ZS & 96.95 & 0.28 & -96.67 & 84.48 & 0.29 & -84.19 & 85.52 & 0.61 & -84.91 \\
qwen2.5-7b-dpo-S & 50.73 & 24.10 & -26.63 & 47.71 & 3.48 & -44.23 & 48.40 & 6.37 & -42.03 \\
qwen2.5-7b-dpo-ZS & 95.32 & 0.37 & -94.95 & 86.23 & 1.17 & -85.06 & 87.75 & 2.35 & -85.40 \\
qwen2.5-7b-sft-S & 0.00 & 30.66 & 30.66 & 3.40 & 3.89 & 0.49 & 6.71 & 6.98 & 0.27 \\
qwen2.5-7b-sft-ZS & 97.17 & 0.00 & -97.17 & 87.91 & 0.25 & -87.66 & 89.51 & 1.26 & -88.25 \\
\bottomrule
\end{tabular}
}
\caption{ \textbf{Complete Performance Analysis on Base and Extended Addition Tasks.} Per-model breakdown of performance (\%) across standard numerical and symbolic representations, with evaluation of degradation ($\Delta$) between formats. Results reveal systematic failures in abstracting arithmetic principles despite high numerical accuracy.}
\label{tab:zero_shot_symbolic_full}
\end{table*}

\end{document}